\documentclass[11pt]{article}
\usepackage[]{acl}

\usepackage{times}
\usepackage{latexsym}
\usepackage{tabularx}
\usepackage{multirow}
\usepackage{enumitem}

\usepackage[T1]{fontenc}
\usepackage[utf8]{inputenc}
\usepackage{microtype}
\usepackage{inconsolata}
\usepackage{graphicx}

\usepackage{booktabs}
\usepackage{placeins} 
\usepackage{array}
\usepackage[table]{xcolor}
\usepackage{amsmath}

\definecolor{darkgreen}{rgb}{0,0.5,0}
\definecolor{darkred}{rgb}{0.6,0.0,0}

\newcolumntype{L}{>{\raggedright\arraybackslash}X}

\usepackage{changes}

\title{Biasless Language Models Learn Unnaturally: How LLMs Fail to Distinguish the Possible from the Impossible}

\author{
  Imry Ziv\textsuperscript{1}, Nur Lan\textsuperscript{2,4}, Emmanuel Chemla\textsuperscript{2,3,4} \\[4pt]
  \textsuperscript{1}Tel Aviv University, \textsuperscript{2}Ecole Normale Supérieure, \\
  \textsuperscript{3}Earth Species Project, \textsuperscript{4}EHESS, PSL University, CNRS \\[4pt]
  \texttt{imryziv@mail.tau.ac.il, \{nur.lan,emmanuel.chemla\}@ens.psl.eu}
}

\begin{document}
\maketitle

\begin{abstract}
Are large language models (LLMs) sensitive to the distinction between humanly possible and impossible languages? This question was recently used in a broader debate on whether LLMs and humans share the same innate learning biases. Previous work has answered it in the positive by comparing LLM learning curves on existing language datasets and on "impossible" datasets derived from them via various perturbation functions. Using the same methodology, we examine this claim on a wider set of languages and impossible perturbations. We find that in most cases, GPT-2 learns each language and its impossible counterpart equally easily, in contrast to previous findings. We also apply a more lenient condition by testing whether GPT-2 provides any kind of separation between the whole sets of natural vs. impossible languages, based on cross-linguistic variance in metrics derived from the learning curves. Taken together, these perspectives show that GPT-2 provides no systematic separation between the possible and the impossible.


\end{abstract}

\section{Introduction}
\subsection{Preliminaries}
A well-known thought experiment in the field of linguistics is that of Chomsky's Martian linguist \citep{Chomsky:2000}. The idea is as follows:  if a Martian linguist were to arrive on Earth, they "might reasonably conclude that there is a single human language, with differences only at the margins" \citep[p.~7]{Chomsky:2000}. Despite the seemingly vast differences between the world's languages, all languages are shaped and restricted by the learning biases that humans are biologically endowed with. 
\begin{figure}[h]
    \centering
    \includegraphics[width=\linewidth]{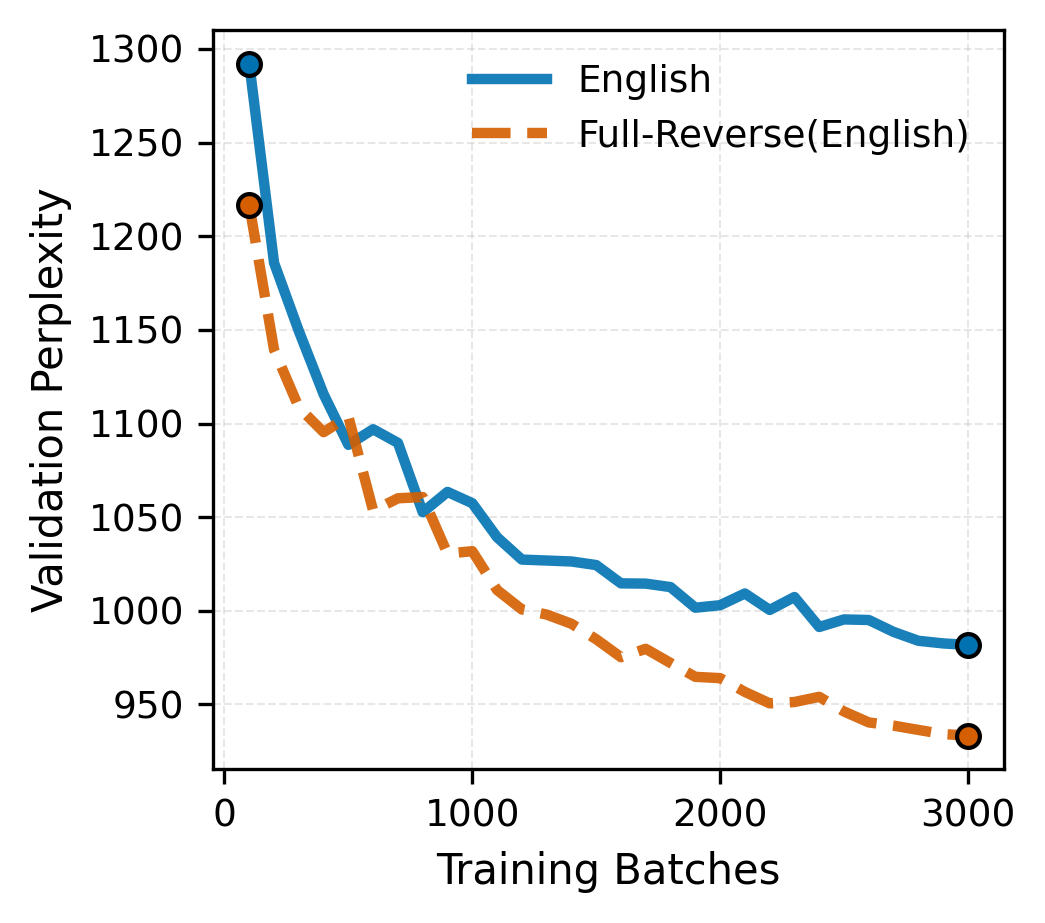}
    \caption{Learning curve of GPT-2 on a standard English dataset vs. a perturbed, impossible variant of the same language. The ease-of-learning methodology used in \citet{KalliniPapadimitriouFutrellMahowaldPotts:2024} fails to find a stable boundary between possible and impossible languages, often preferring the impossible variants as shown here. See Figure~\ref{fig:language-learning-curves} for the full results.}
    \label{fig:demo-english-full-reverse}
\end{figure}
In this spirit, linguists have discovered patterns which seem to hold across virtually all known natural languages, seeking to uncover what the Martian metaphor has alluded to: the restrictions that define what a human language may potentially be \citep{Baker:2012}. One example of such a pattern is the fact that dependencies in natural language are universally stated in terms of hierarchical representations and never in terms of linear representations \citep{Chomsky:1965}. At the same time, linguists also find that some patterns are consistently absent from all attested languages. That is the case of so-called linguistic "islands" \citep{Ross:1967}, restriction conditions on syntactic movement which apply across virtually every known language. These varied patterns reveal a highly skewed language typology: one in which learners are drawn to very specific kinds of patterns while staying oblivious to otherwise-plausible ones.

Based on this empirical evidence, one would arrive at what we name \textit{the generative hypothesis}: that the set of humanly attested languages is actually a small subset of all potentially observable languages, and that key aspects of linguistic typology are shaped this way because of strong learning biases humans are innately endowed with.\footnote{\label{fn: not-all-typological}Not all typological asymmetries stem from the learning biases --- they could arise from many functional pressures, such as communicative pressures \citep{FutrellHahn:2022} or  historical trajectories \citep{Collins:2016}. We make sure to test only for asymmetries related to the human language device, see Section~\ref{sec:impossible-unnatural-unattested}.}
Hypothetical languages that do not adhere to these universal constraints came to be known as \textit{impossible languages}. Some have argued that in humans, the notion of impossible languages is biologically embedded, evidenced by selective functional recruitment of different brain regions for the processing of possible languages and impossible ones (\citealp{Musso:2003}, \citealp{Moro:2016}, \citealp{MoroGrecoCappa:2023}). 

The recent advent of large language models (LLMs) and their subsequent use as cognitive modeling tools (see \citealp{Futrell:2025} for an extensive review) has sparked a similar question regarding such models: are LLMs sensitive to the distinction between possible and impossible languages? Throughout this article we will refer to this issue as \textit{sensitivity to impossibility}. As previously noticed (\citealp{chomsky2023falsepromise}, \citealp{MoroGrecoCappa:2023}, \citealp{ZivLanChemlaKatzir:2025}), a negative answer to this question may undermine LLMs as cognitive models for human linguistic cognition, since this distinction is thought to be shaped by human learning biases. On the other hand, a positive answer would provide evidence against \textit{the generative hypothesis}.

In the following sections, we provide experimental support for a negative answer to this question. We build on the methodology from \citet{KalliniPapadimitriouFutrellMahowaldPotts:2024}, which measures ֿ\textit{sensitivity to impossibility} by comparing the validation perplexities of GPT-2 models when trained on English datasets and on artificially perturbed versions of these datasets. The perturbation functions generate linguistically impossible variants of existing datasets mainly by performing sentence-level shuffles and reversals (a common method for creating "unnatural" datasets, see \citealp{MitchellBowers:2020}, and \citealp{Sinha:2021}, among others). While \citet{KalliniPapadimitriouFutrellMahowaldPotts:2024} provide a positive answer to the question of LLM \textit{sensitivity to impossibility}, we extend their methodology to additional languages and perturbations, and show that LLMs are equally capable of learning possible languages and their perturbed impossible counterparts, in some cases showing a strong preference for the impossible variants. The cross-linguistic methodology allows us to also examine whether the LLM provides any kind of separation between the whole set of possible languages and the whole set of impossible languages. To inform this perspective, we compute basic metrics (minimal perplexity value achieved during training, area under perplexity curve) that reflect the ease-of-learning of the language in question, and then compare the variation in these metrics between possible languages with the variation between the different impossible variants of each language. Here, too, the \citet{KalliniPapadimitriouFutrellMahowaldPotts:2024} method fails to generalize cross-linguistically, as we find that each possible language patterns with its impossible variants. We conclude that if the perplexity method is indeed a good proxy for ease-of-learning by the LLM, then LLMs do not share the human learning biases that shape linguistic typology and render the perturbed languages impossible.

\subsection{Impossible, Unnatural, and Unattested}
\label{sec:impossible-unnatural-unattested}
We note that there has been some ambiguity in the literature regarding the terms \textit{impossible languages}, \textit{unnatural languages}, and \textit{unattested languages}. As pointed out by \citet{Xu:2025}, there is an important difference between languages that are humanly impossible and languages that are implausible, where the latter merely violate typological tendencies; not every implausible language can meaningfully bear on cognitive issues (one can imagine an unattested version of English in which every sentence is concatenated to itself four times as an absurd example, and see also Footnote~\ref{fn: not-all-typological}).  

In this work, we will use the term \textit{impossible languages} to refer to hypothetical languages that violate innate biases that are specifically linguistic. This is important since there exist certain typological tendencies that have been attributed to the workings of domain-general learning mechanisms (for example, the prevalence of harmonic word orders, see \citealp{Culbertson:2015}), or to the effect of information-theoretical channel biases that guide language evolution, thus shaping the typological landscape (\citealp{Clark2023}, \citealp{FutrellHahn:2025}). These are the kinds of typologically marked phenomena that have been at the center of work similar but orthogonal to ours, such as \citet{Xu:2025}. In contrast, the main linguistic bias our work manipulates is the reliance on recursive procedures to generate hierarchical structures, which is considered at the core of human linguistic cognition (\citealp{Chomsky:1965}, \citealp{HauserChomskyFitch:2002}). Weaker typological tendencies do not fit into an argument regarding how learning biases in the language faculty shape the typology. 

\section{Related Work}
\label{sec:related-work}
\subsection{The Cognitive Relevance of Large Language Models}

The impressive performance of deep learning models on many linguistic benchmarks has fueled growing interest in employing artificial neural networks to model human linguistic cognition (e.g., \citealp{LinzenBaroni:2020}, \citealp{Baroni:2022}, \citealp{WilcoxFutrellLevy:2023}, \citealp{Futrell:2025}). This kind of work usually relies on the assumption that artificial neural networks are proxies for undisclosed theories of human linguistic cognition \citep{ZivLanChemlaKatzir:2025}. These proxies are relatively unbiased, in the sense that
they were not built with the kinds of biases that linguistics has argued for in order to explain humans' learning processes and outcomes. If the performance of such proxies matches that of humans on key linguistic phenomena, one can argue that this weakens \textit{the generative hypothesis} (although the claim that artificial neural networks are bias-free has been contested, see \citealp{ChemlaNefdt:2024}). This has been the argumentation in works such as \citet{WilcoxFutrellLevy:2023}, \citet{MahowaldIvanovaBlankKanwisherTenenbaumFedorenko:2024}, and \citet{KalliniPapadimitriouFutrellMahowaldPotts:2024}. 

There are other objections to the use of LLMs as models of human linguistic cognition: \citet{FoxKatzir:2024} show that LLMs fail to adhere to elementary observations in linguistics, such as the distinction between competence and performance \citep{Yngve:1960} and between correctness and likelihood \citep{Chomsky:1957}. LLMs are also inconsistent with human learning and processing of certain linguistic phenomena such as syntactic islands \citep{LanChemlaKatzir:2024a}. These findings imply that LLMs may fail to capture patterns that humans learn well, reducing their value as models of human cognition. 
Here we check whether the LLM proxy evaluated in \citet{KalliniPapadimitriouFutrellMahowaldPotts:2024} remains robust under further scrutiny on additional languages, perturbations, and statistical tests.

\subsection{Large Language Models and Impossible Languages}
A particularly interesting line of work within the use of LLMs as models of human linguistic cognition argues that large language models exhibit human-like sensitivity to the distinction between possible and impossible languages. Continuing a line of inquiry concerning the performance of statistical parsers on unnatural language constructions (\citealp{FongBerwick:2008}, \citealp{Fong:2013}), \citet{MitchellBowers:2020} have shown that recurrent neural networks (RNNs, \citealp{Elman:1990}) learn number agreement easily even within impossible constructions such as partially reversed sentences and randomly shuffled vocabularies. Later work (e.g., \citealp{Abdou:2022}) examines the effect of shuffling and disrupting fixed word orders on the performance of Transformer language models \citep{Vaswani:2017} such as BERT \citep{Devlin:2018}.

More recently, \citet{KalliniPapadimitriouFutrellMahowaldPotts:2024} address \textit{sensitivity to impossibility} using an approach we adopt in the current paper: they attempt to show that LLMs' learning curves during training reflect a learning preference for English over languages that were generated from English datasets via various "unnatural" perturbations. 
It was shown in that work that GPT-2's learning curve (in terms of perplexity on a held-out corpus) when trained on an English dataset remains constantly below the same curve for unattested, artificially-generated variants of English. The unattested languages are generated using perturbation functions that deliberately disrupt tendencies thought to be universal --- for example, by reversing or shuffling elements in ways that break constituent structure and disrupt word order.
\citet{KalliniPapadimitriouFutrellMahowaldPotts:2024} took these results to suggest that asymmetries in LLMs' learning curves can account for typological asymmetries -- here, the non-existence of certain languages and the prevalence of others -- and that by doing so they help discard linguistic biases. \citet{yang2025anything} make a similar case while extending this methodology to additional languages, and \citet{Xu:2025} consider perturbations such as counterfactual word orders, that are not impossible but typologically implausible.

If \citeposs{KalliniPapadimitriouFutrellMahowaldPotts:2024} answer to the \textit{sensitivity to impossibility} issue proves true, it would undermine \textit{the generative hypothesis}, considering that LLMs lack the learning biases that humans come equipped with. As will become clearer in the following sections, our work adds to this debate by providing evidence to the contrary: GPT-2 fails to distinguish the possible from the impossible across many novel perturbations and languages, both interlinguistically and intralinguistically.

\subsection{The Intralinguistic and Interlinguistic Perspectives}
Using LLM learning curves as proxies for \textit{sensitivity to impossibility} raises two kinds of questions. First, are LLMs sensitive to the distinction between a language and its impossible counterpart? We dub this question the \textit{intralingustic perspective}, and note that this is the perspective that was examined in \citet{KalliniPapadimitriouFutrellMahowaldPotts:2024}, although on English only. The second question pertains to an \textit{interlinguistic perspective}: are LLMs sensitive to the distinction between the set of all attested languages and the set of all unattested ones? One might argue that if LLM learning curves provide an adequate explanation for the typological landscape, this should be reflected in all attested languages' learning curves remaining below all unattested ones' (similarly to recent work by \citealp{yang2025anything}). A key objection to this requirement is that language typology is shaped by factors other than the implicit biases of the learner. To account for this, we test instead a weaker criterion that controls for external factors that might affect the typology: whether the variance in learning curves between attested languages and their unattested counterparts is greater than the variance among attested languages.

\section{Methodology}

\subsection{Impossible Perturbations}
\label{impossible-perturbations}
We test the two perspectives based on the methodology by \citet{KalliniPapadimitriouFutrellMahowaldPotts:2024}. For each attested language (e.g., Italian) we create perturbed versions of that language which are impossible -- for example, a version of Italian in which a syntactic transformation relies on linear word position (e.g., reversing a sentence from the fourth token onwards). The perturbations are listed in Table~\ref{table:perturbations}.
Each language (either possible or impossible) then serves as a training dataset for a GPT-2 model \cite{Radford:2019}.

\label{sec:methodology}

\begin{table}[h!]
    \centering
    \footnotesize  
    \renewcommand{\arraystretch}{1.5}  
    \resizebox{0.5\textwidth}{!}{  
    \begin{tabular}{p{1.7cm}p{5cm}}  
        \toprule
        \textbf{Perturbation} & \textbf{\textcolor{blue}{Baseline} / \textcolor{orange}{*Perturbed}} \\ 
        \midrule
        \texttt{SHUFFLE global} &
        \parbox[t]{5cm}{\textcolor{blue}{Colorless\textsubscript{0} green\textsubscript{1} ideas\textsubscript{2} sleep\textsubscript{3} furiously\textsubscript{4.}}\\
                        \textcolor{orange}{*Sleep\textsubscript{3} ideas\textsubscript{2} colorless\textsubscript{0} furiously\textsubscript{4} green\textsubscript{1}}} \\
        \midrule
        \texttt{SHUFFLE local} &
        \parbox[t]{5cm}{\textcolor{blue}{Colorless\textsubscript{0} green\textsubscript{1} ideas\textsubscript{2} sleep\textsubscript{3} furiously\textsubscript{4.}}\\
                        \textcolor{orange}{Green\textsubscript{1} colorless\textsubscript{0} sleep\textsubscript{3} ideas\textsubscript{2} furiously\textsubscript{4.}}} \\
        \midrule
        \texttt{REVERSE partial} & 
        \parbox[t]{5cm}{\textcolor{blue}{Colorless\textsubscript{0} green\textsubscript{1} \texttt{<rev>}\textsubscript{2} ideas\textsubscript{3} sleep\textsubscript{4} furiously\textsubscript{5.}}\\
                        \textcolor{orange}{*Colorless\textsubscript{0} green\textsubscript{1} \texttt{<rev>}\textsubscript{2} furiously\textsubscript{5} sleep\textsubscript{4} ideas\textsubscript{3}.}} \\
        \midrule
        \texttt{REVERSE \linebreak full} & 
        \parbox[t]{5cm}{\textcolor{blue}{Colorless\textsubscript{0} green\textsubscript{1} \texttt{<rev>}\textsubscript{2} ideas\textsubscript{3} sleep\textsubscript{4} furiously\textsubscript{5.}}\\
                        \textcolor{orange}{*Furiously\textsubscript{5} sleep\textsubscript{4} ideas\textsubscript{3} \texttt{<rev>}\textsubscript{2} green\textsubscript{1} colorless\textsubscript{0}.}} \\
        \midrule
        \texttt{SWITCH} & 
        \parbox[t]{5cm}{\textcolor{blue}{Colorless\textsubscript{0} green\textsubscript{1} ideas\textsubscript{2} sleep\textsubscript{3} furiously\textsubscript{4}.}\\
                        \textcolor{orange}{*Ideas\textsubscript{2} green\textsubscript{1} colorless\textsubscript{0} sleep\textsubscript{3} furiously\textsubscript{4}.}} \\
        \midrule
        \texttt{\texttt{HOP}} & 
        \parbox[t]{5cm}{\textcolor{blue}{They\textsubscript{0} were\textsubscript{1} sleeping\textsubscript{2} \textbf{v\textsubscript{3}} next\textsubscript{4} to\textsubscript{5} the\textsubscript{6} colorless\textsubscript{7} green\textsubscript{8} ideas\textsubscript{9}.}\\
                        \textcolor{orange}{*They\textsubscript{0} were\textsubscript{1} sleeping\textsubscript{2} next\textsubscript{3} to\textsubscript{4} the\textsubscript{5} \textbf{v\textsubscript{6}} colorless\textsubscript{7} green\textsubscript{8} ideas\textsubscript{9}.}} \\
        \bottomrule
    \end{tabular}
    }
    \caption{Impossible perturbations used in our experiments. The perturbations are applied to attested languages to generate impossible variants of them.}
    \label{table:perturbations}
\end{table}

\begin{figure*}[h!]
    \centering
    \includegraphics[width=1.0\textwidth]{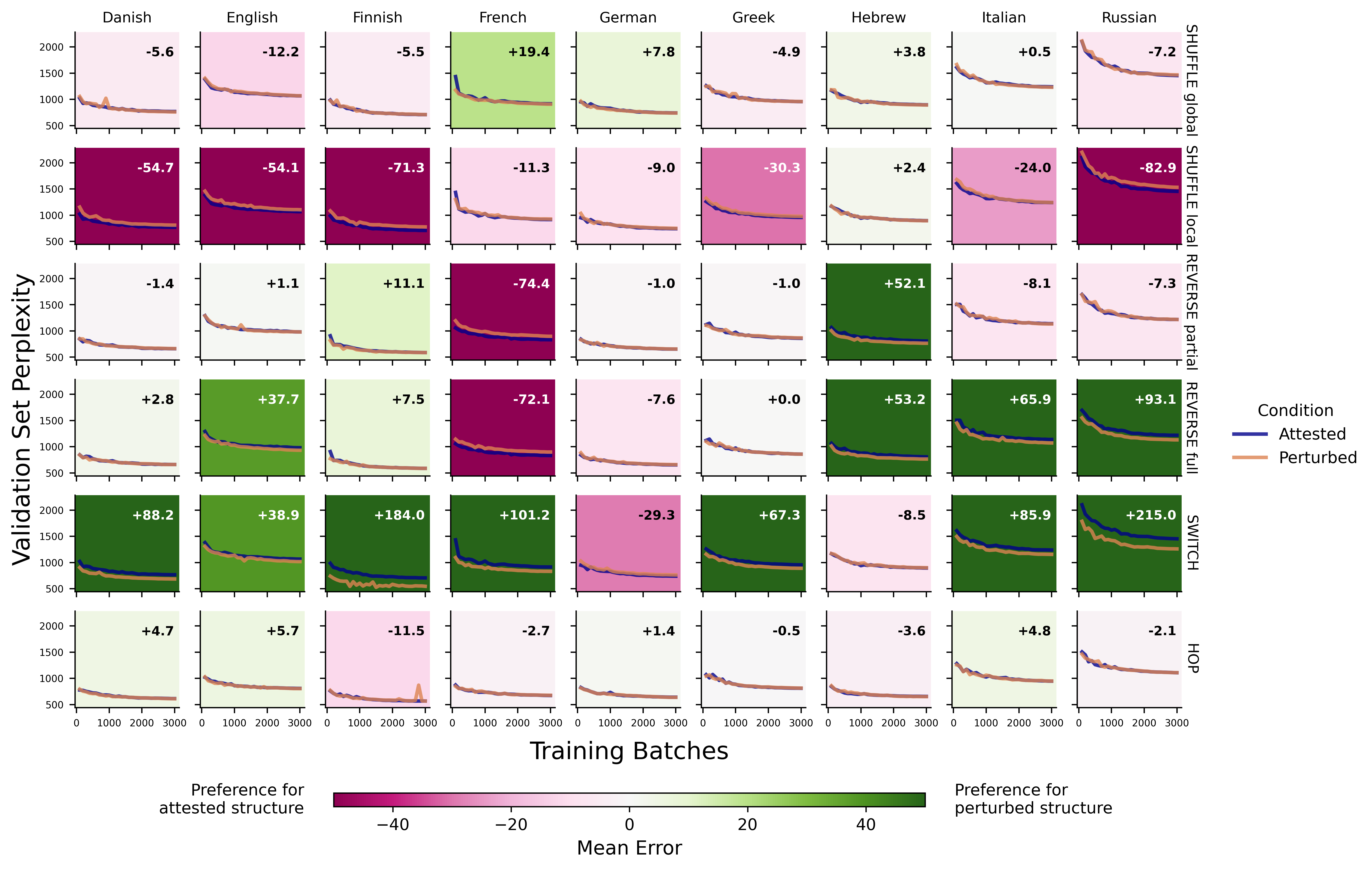}
    \caption{Learning curves for attested languages (dark blue curves) and their perturbed variants (orange). Each subplot displays the learning curves for an experiment alongside mean error values, representing the mean difference between the two learning curves. Positive values (green hues) indicate that the attested language's learning curve is, on average, above its perturbed variant, while negative values (pink hues) indicate the opposite.}
    \label{fig:language-learning-curves}
\end{figure*}

We expand the experimental coverage in \citet{KalliniPapadimitriouFutrellMahowaldPotts:2024} in two ways. First, while \citet{KalliniPapadimitriouFutrellMahowaldPotts:2024} only trained on English (and impossible variants of it), we add eight additional languages: Danish, Finnish, French, German, Greek, Hebrew, Italian, and Russian. 

Second, for each attested language we apply the perturbations given in Table~\ref{table:perturbations}. \texttt{SHUFFLE} perturbations change the word order in a sentence: \texttt{SHUFFLE global} shuffles each sentence deterministically based on sentence length in tokens, and \texttt{SHUFFLE local} switches each even-indexed token with the following odd-indexed token. The perturbation \texttt{SWITCH} swaps the tokens at indices 0 and 2 (\texttt{SWITCH} and \texttt{SHUFFLE} perturbations are compared to the baseline dataset \texttt{NO PERTURB}). 

\texttt{REVERSE} perturbations flip the word order in a sentence: \texttt{REVERSE full} reverses the entire sentence while \texttt{REVERSE partial} reverses the order starting after a randomly inserted \texttt{<rev>} marker. \texttt{REVERSE} languages are compared to a baseline dataset \texttt{REVERSE baseline} where the \texttt{<rev>} token is inserted at the same location without additional changes, to control for the effect of additional textual material on perplexities. 

\texttt{\texttt{HOP}} inserts a special marker three tokens after each verb. \texttt{\texttt{HOP}} is compared to a baseline \texttt{\texttt{HOP} baseline} where the same marker is inserted right after the verb, again to control for perplexity effects of the marker. 

Note that almost all employed perturbations disrupt the basic notion of hierarchy in natural language: The \texttt{SHUFFLE}, \texttt{SWITCH}, and \texttt{REVERSE partial} perturbations ignore hierarchical structure completely and break up constituent structure. While not directly at odds with hierarchical structure, the \texttt{REVERSE full} perturbation violates universal linguistic notions such as binding and anaphora \citep{Chomsky1981}, and the \texttt{\texttt{HOP}} perturbation requires a linear generalization (i.e., counting tokens sequentially) to be learned.

\subsection{Baseline Dataset Creation}
\label{sec-baseline-creation}
\begin{figure*}[h!]
    \centering
    \makebox[\textwidth][c]{%
        \includegraphics[width=1.0\textwidth]{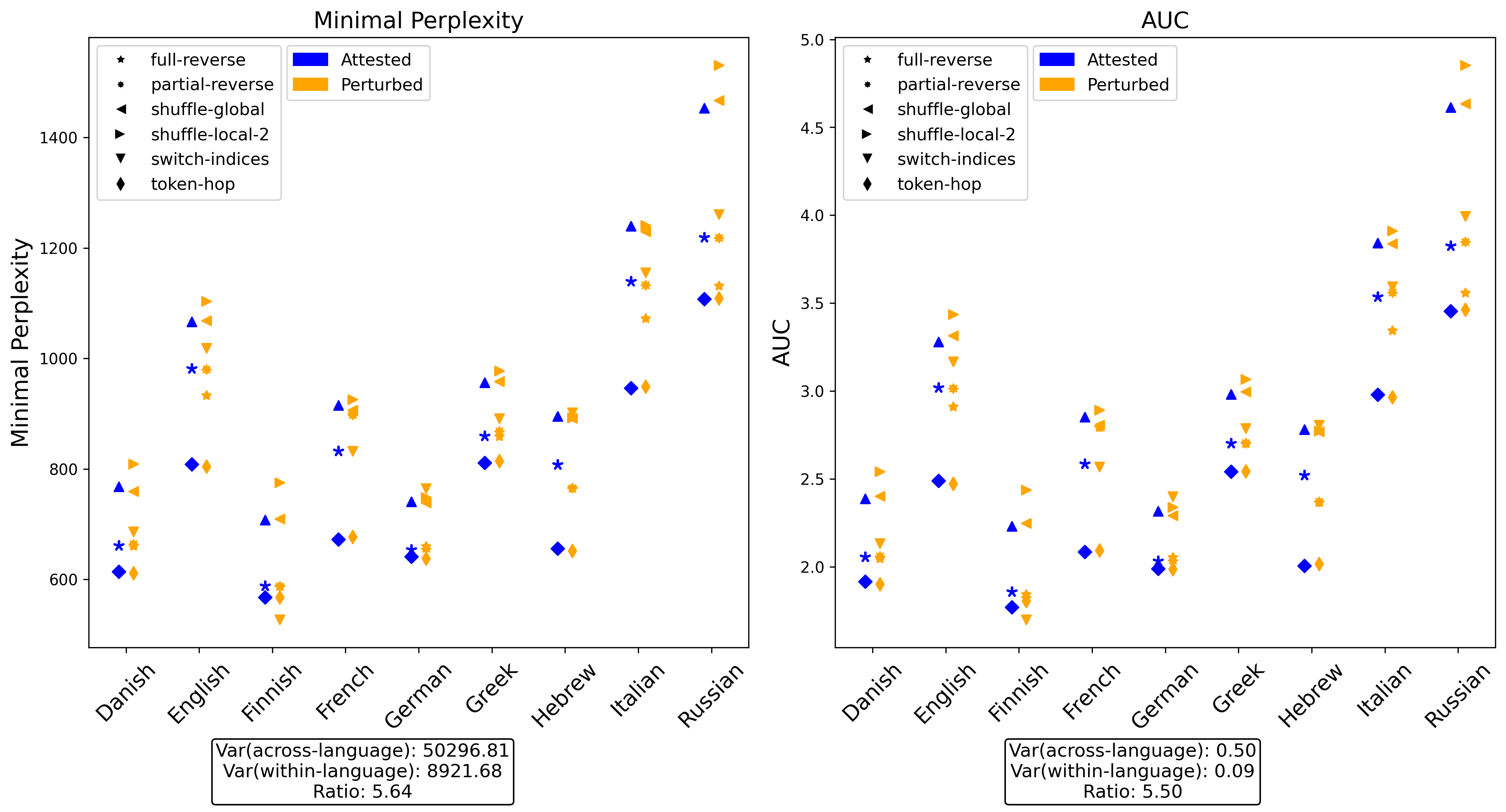}%
    }
    \caption{Cross-linguistic comparison of minimal validation perplexity values during training and of area under the curve (AUC) of the training curves from Figure~\ref{fig:language-learning-curves}. Languages compared to the same baseline have the same shape: triangles are compared to \texttt{NO PERTURB} (blue triangle), stars to \texttt{REVERSE baseline} (blue star) and diamonds to \texttt{\texttt{HOP} baseline} (blue diamond). \textit{across-language} and \textit{within-language} variances are denoted below each subplot. Values vary considerably more across languages (different languages across same perturbation) than within languages (different perturbations across same language). 
    This shows that attested languages and their perturbed, impossible variants pattern together.}
    \label{fig:cross-linguistic-metrics}
\end{figure*}

To create our baseline datasets, we replicate the dataset creation procedure used in \citet{GulordavaBojanowskiGraveLinzenBaroni:2018}. We extract recent Wikipedia dumps using \hyperlink{https://github.com/attardi/wikiextractor}{WikiExtractor}, clean them with TreeTagger \citep{Schmid:1999} and tokenize them using a simple whitespace tokenizer, uniform for all languages. We then consider 90M token subsets shuffled at sentence level as our final baseline dataset. We set a uniform tokenizer vocabulary size of 50,257 per language, following work such as \citet{ArnettBergen:2025}. Most dumps used were from January 2025, except English, Italian, Russian, and Hebrew for which we used the original datasets created by \citet{GulordavaBojanowskiGraveLinzenBaroni:2018}. Our uniform dataset creation method ensures that the differences in perplexities stem only from inherent properties of the languages we consider, and not from unrelated variables such as non-uniform tokenization, sentence lengths, number of unknown tokens, different language registers due to different sources, etc. The languages used were chosen based on the following factors: whether they have a recent Wikipedia dump of sufficient size ($\geq$100M tokens) and whether they have a publicly available part-of-speech (POS) tagger with $\geq$90\% accuracy. Alongside these considerations, our choice of languages was also informed by trying to choose languages from distinct language branches. Our choices yield seven distinct branches: North Germanic (Danish), West Germanic (English, German), Romance (French, Italian), Hellenic (Greek), Uralic-Finnic (Finnish), Semitic (Hebrew) and East Slavic (Russian).

\subsection{Impossible Dataset Creation}

After creating a \textit{train}, \textit{test}, \textit{validation} split for each baseline dataset, we run our perturbation functions (see Table~\ref{table:perturbations}) on each split, sentence by sentence, similarly to~\citet{KalliniPapadimitriouFutrellMahowaldPotts:2024}. To make sure that each sentence of each perturbed dataset has the same number of tokens as its corresponding sentence in the baseline dataset, we insert additional baseline tokens in the baseline variants of perturbations that require additional text (\texttt{REVERSE partial}, \texttt{REVERSE full}, \texttt{\texttt{HOP}}, see Table~\ref{table:perturbations}). As mentioned, this controls for the perplexity effect of the insertion of the \texttt{REVERSE/\texttt{HOP}} markers. 

Our \texttt{\texttt{HOP}} perturbations require POS tagging, which we mostly performed using the \href{https://spaCy.io/models}{spaCy} Python library (see Table~\ref{tab:spaCy}). For each language we choose a \texttt{\texttt{HOP}} marker that is a single-character token that did not previously exist in the model's vocabulary. We choose to replace \citet{KalliniPapadimitriouFutrellMahowaldPotts:2024}'s 3rd person agreement \textit{token-hop} with POS-based marking because of the lower accuracy of morphological taggers on languages that are not English and cross-linguistic differences in agreement morphology.

\subsection{Model Training}
We train a GPT-2 model from scratch on the tokenized corpora using HuggingFace's Transformers' \textit{GPT2LMHeadModel}, with a standard configuration: 12 layers, 12 attention heads, 768-dimensional embeddings, and a context size of 1024 tokens. Following \citet{KalliniPapadimitriouFutrellMahowaldPotts:2024}, training uses batch size 512, a learning rate of $5 \times 10^{-4}$, and weight decay of $10^{-2}$, 
stopping after 3,000 batches. We record validation set perplexity every 100 batches. Performance is compared by plotting learning curves for each possible/impossible pair. Training is performed using the default Trainer module from the HuggingFace \textit{transformers} Python package. The experiments were run on Nvidia V100 and A40 GPUs. Each perturbation experiment lasted $\sim1$ hour, amounting to $\sim81$ GPU hours for all languages and perturbations.

\subsection{Operationalizing \textit{sensitivity to impossibility}}

To test the \textit{intralinguistic perspective}, we plot the perplexity curves for each attested language and its perturbed impossible counterpart. For each such pair we compute mean error (ME):
\[
\frac{1}{N} \sum_{i=1}^{N} (B_i - M_i)
\]
where $B_i$ are points on the baseline/possible curve, $M_i$ are points on the perturbed/impossible curve, and $N$ is the number of perplexity measurement points. A positive ME is indicative of model failure in explaining the intralinguistic perspective: it means that the baseline curve is consistently above --- or coincides with --- the perturbed, impossible curve.

To test the \textit{interlinguistic perspective}, we compare the minimum perplexity value achieved during training as well as the area under the curve (AUC) for all possible and impossible languages. We consider the variance in these metrics between possible languages (across-language variance) and between the different variants of each language (within-language variance).\footnote{This operationalization of separability is different from the linear SVM classifier method used in \citet{yang2025anything}.} If within-language variance is significantly lower than across-language variance, LLMs' learning curves cannot account for typological asymmetries, and the question of \textit{sensitivity to impossibility} in LLMs receives a negative answer.\footnote{All experimental materials, source code, and results are available at \url{https://github.com/imry-ziv/impossible-languages-acl}.}


\section{Results}

\subsection{Intralinguistic Perspective}

The learning curves for the different languages are shown in Figure~\ref{fig:language-learning-curves}. We plot the learning curve for each possible language alongside its impossible variants. In an overwhelming amount of languages, the learning curves coincide almost completely. In cases such as \texttt{\{english, italian\}-REVERSE full} and all \texttt{SWITCH} languages except Hebrew and German, the impossible language's learning curve remains below its baseline. 
Figure~\ref{fig:language-learning-curves} also shows that mean error is often positive (green), indicating the model's preference for the impossible structure. Out of 54 test cases, 28 are in the expected direction, for which a negative mean error indicates that the possible curve is on average below its impossible counterpart. A two-sided binomial test comparing this result to chance ($P$ = 0.5) yields a $p$-value of 0.892, indicating no
statistically significant deviation from random expectation. The perturbation best explained by the model is \texttt{SHUFFLE local}, with 8 out of 9 languages in the right direction, and the worst ones are \texttt{REVERSE full} and \texttt{SWITCH}, in which the model fails to prefer the attested variant for 7 out of 9 languages. Since GPT-2 does not prefer possible languages across language pairs in a statistically significant manner, it fails to explain \textit{the intralinguistic perspective}.

\subsection{Interlinguistic Perspective}
Figure~\ref{fig:cross-linguistic-metrics} plots the minimal perplexity and AUC values for each of the experiments. On the interlinguistic perspective, it is apparent that possible languages and their impossible counterparts pattern together, with a distinct cluster for each attested language and all its perturbations. Russian, Italian, and all their perturbations, for example, are projected as harder than all impossible variants of all other languages. GPT-2 provides no clear-cut separation in these metrics between the set of humanly attested languages (blue data points) and the set of impossibly perturbed ones (orange data points). 

Regarding the variances, \textit{across-language} variance compares different languages under the same perturbation, capturing differences in ease-of-learning between languages. \textit{Within-language} variance, by contrast, compares different perturbations of the same language, capturing differences between an attested language and its perturbed counterparts. If the model were to correctly separate the possible from the impossible, we would expect higher \textit{within-language} variances: attested languages should pattern together, while impossible variants should be projected as significantly harder, thereby increasing \textit{within-language} variance. 

Our results demonstrate the opposite (see Figure~\ref{fig:cross-linguistic-metrics}). This is not expected if relative ease-of-learning, taken as the relative ordering of learning curves and measured by minimum perplexity and AUC, is an adequate predictor of attested typologies. GPT-2's failure to explain both perspectives suggests that it does not share the human innate biases that shape linguistic typology according to \textit{the generative hypothesis}, and provide no evidence against strong innate biases in human linguistic cognition.

\section{Discussion}
\label{sec:disc}
The rise of LLMs, first within the field of natural language processing (NLP) and later in multiple areas of science, has offered a helpful new tool with which one can revisit questions raised by linguists since the 1950's. Recent literature (\citealp{KalliniPapadimitriouFutrellMahowaldPotts:2024}, \citealp{yang2025anything}, \citealp{Xu:2025}) has suggested that LLMs show that observations about language typology can be explained without assuming strong linguistic biases, namely, by examining the \textit{sensitivity to impossibility} of Transformer-based language models such as GPT-2. If these models show such sensitivity, it would weaken the position that humans are born with strong innate learning biases, and that these biases are what shapes the possible-impossible frontier in linguistic typology (\textit{the generative hypothesis}). 

Here we provide an empirical examination of the claim that shows that so-called ease-of-learning, measured through perplexity curves over training, fails to tease apart attested languages and versions of them that are modified in humanly impossible ways. In most pairs of languages, the LLMs' learning curves actually indicated a dispreference (or indifference) towards existing languages when compared to impossible ones, contrasting with previous results from \citet{KalliniPapadimitriouFutrellMahowaldPotts:2024} and \citet{yang2025anything}. Ease-of-learning also fails to create a stable boundary between attested and unattested languages, as exhibited by the interlinguistic perspective in Figure~\ref{fig:cross-linguistic-metrics}. Learning curves thus seem to be an irrelevant measure for deriving language typology, at least for the question of language \mbox{(non-)existence}. This result holds across the variety of languages we tested. Again, this suggests that the biases present in the model architecture and/or training regimes of LLMs are quite different from those that human learners are equipped with.\footnote{A similar stance contesting the \citet{KalliniPapadimitriouFutrellMahowaldPotts:2024} approach has been taken in recent work, both conceptually (\citealp{Milway:2025}, \citealp{Hunter:2025}) and empirically \citep{Leivada:2025}.}

\section{Limitations}

While we expand coverage relative to previous work, our experiments rely on a limited number of perturbations and languages, which cannot capture the full diversity of the typological landscape. Future work could apply the same methodologies to a larger set of linguistic phenomena and perturbations to strengthen our conclusions.

Moreover, the perplexity curve metric used in our experiments does not suffice as a complete explanation of typological asymmetries, as there are many other factors that shape the typology other than learning asymmetries and innate biases. Future work could embed the learning component within a model of cultural evolution, in which small learning asymmetries may be amplified across generations (see \citealp{KirbyDowmanGriffiths:2007}, \citealp{NiyogiBerwick:2009}, and \citealp{BrochhagenFrankeRooij:2018}, among others).

Another limitation is that our experiments rely on GPT-2. This choice was made to allow for direct comparison with the methodology in \citet{KalliniPapadimitriouFutrellMahowaldPotts:2024} who relied on GPT-2 alone as well.
It is possible that the results would not extend to state-of-the-art LLMs. Future work could investigate whether either conclusion holds for larger-scale models.

We also note that the current work differs from previous work on several technical aspects which might explain our different results. First, while \citet{KalliniPapadimitriouFutrellMahowaldPotts:2024} and \citet{yang2025anything} used pretrained tokenizers for each language, we use a uniform whitespace tokenizer with a vocabulary size of 50,257 for all languages. This was meant to unify the different learning tasks.
This might have disadvantaged the learning of more morphologically-complex languages, but we do not expect it to affect the results within same-language comparisons. Our setup also differs in hyperparameter, training set size, and training regime choices (e.g., we do not use the learning rate warmup used in \citealp{KalliniPapadimitriouFutrellMahowaldPotts:2024}). Admittedly these choices could tip the learning curve patterns in the opposite direction. We note however that as long as the language modeling setup is reasonable, the cognitive linking hypothesis (i.e., the ease-of-learning methodology based on learning curves) should not be so fragile as to break based on hyperparameter choices.




\section*{Acknowledgments}

The authors would like to thank the TAU Computational Linguistics Lab, the MIT Computational Psycholinguistics Lab, the LINGUAE research group at ENS, and the anonymous ACL reviewers for their helpful comments and discussions. This project was provided with computer and storage resources by GENCI at IDRIS thanks to the grant AD011013783R2 on the supercomputer Jean Zay's V100 and CSL partitions.

\bibliography{jlm}

@misc{ZivLanChemlaKatzir:2025,
  title = {Large {{Language Models}} as {{Proxies}} for {{Theories}} of {{Human Linguistic Cognition}}},
  author = {Ziv, Imry and Lan, Nur and Chemla, Emmanuel and Katzir, Roni},
  year = {2025},
  month = feb,
  number = {arXiv:2502.07687},
  eprint = {2502.07687},
  primaryclass = {cs},
  publisher = {arXiv},
  doi = {10.48550/arXiv.2502.07687},
  urldate = {2025-02-27},
  archiveprefix = {arXiv}
}

@inproceedings{MitchellBowers:2020,
    title = "Priorless Recurrent Networks Learn Curiously",
    author = "Mitchell, Jeff and Bowers, Jeffrey",
    booktitle = "Proceedings of the 28th International Conference on Computational Linguistics",
    year = "2020"
}

@article{radford:2019,
  author = {Radford, Alec and Wu, Jeffrey and Child, Rewon and Luan, David and Amodei, Dario and Sutskever, Ilya},
  journal = {OpenAI},
  note = {Accessed: 2024-11-15},
  title = {Language Models are Unsupervised Multitask Learners},
  year = 2019
}

@book{Chomsky:1957,
	address = {The Hague},
	author = {Chomsky, Noam},
	publisher = {Mouton},
	title = {Syntactic Structures},
	year = {1957}}

@book{Chomsky:1965,
	address = {Cambridge, MA},
	author = {Chomsky, Noam},
	publisher = {MIT Press},
	title = {Aspects of the Theory of Syntax},
	year = {1965}}

@phdthesis{Ross:1967,
	address = {Cambridge, MA},
	author = {Ross, John R.},
	school = {MIT},
	title = {Constraints on Variables in Syntax},
	year = {1967}}

@article{BrochhagenFrankeRooij:2018,
	author = {Brochhagen, Thomas and Franke, Michael and van Rooij, Robert},
	journal = {Cognitive Science},
	month = {2021/05/21},
	number = {8},
	pages = {2757--2789},
	title = {Coevolution of Lexical Meaning and Pragmatic Use},
	volume = {42},
	year = {2018}}

@article{KirbyDowmanGriffiths:2007,
	author = {Kirby, Simon and Dowman, Mike and Griffiths, Thomas L.},
	journal = {Proceedings of the National Academy of Sciences},
	number = {12},
	pages = {5241-5245},
	title = {Innateness and culture in the evolution of language},
	volume = {104},
	year = {2007}}

@article{NiyogiBerwick:2009,
	author = {Niyogi, Partha and Berwick, Robert C.},
	journal = {Proceedings of the National Academy of Sciences},
	pages = {10124--10129},
	title = {The proper treatment of language acquisition and change in a population setting},
	volume = {106},
	year = {2009}}

@article{Hunter:2025,
	author = {Hunter, Tim},
	journal = {Computational Linguistics},
	month = {8/19/2025},
	number = {2},
	pages = {641--650},
	title = {Kallini et al. (2024) Do Not Compare Impossible Languages with Constituency-based Ones},
	volume = {51},
	year = {2025}}

@article{chomsky2023falsepromise,
  author    = {Noam Chomsky and Ian Roberts and Jeffrey Watumull},
  title     = {Noam Chomsky: The False Promise of ChatGPT},
  journal   = {The New York Times},
  year      = {2023},
  url       = {https://www.nytimes.com/2023/03/08/opinion/noam-chomsky-chatgpt-ai.html}, 
  note      = {Accessed: 2025-09-10} 
}

@article{WilcoxFutrellLevy:2023,
	author = {Wilcox, Ethan Gotlieb and Futrell, Richard and Levy, Roger},
	journal = {Linguistic Inquiry},
	month = {1/13/2024},
	pages = {1--44},
	title = {Using Computational Models to Test Syntactic Learnability},
	year = {2023},
      doi = {10.1162/ling_a_00491}
}

@article{MoroGrecoCappa:2023,
	author = {Moro, Andrea and Greco, Matteo and Cappa, Stefano F.},
	journal = {Cortex},
	pages = {82--85},
        doi = {https://doi.org/10.1016/j.cortex.2023.07.003},
        title = {Large languages, impossible languages and human brains},
	volume = {167},
	year = {2023}}

@article{LanChemlaKatzir:2024a,
	author = {Lan, Nur and Chemla, Emmanuel and Katzir, Roni},
	journal = {Linguistic Inquiry},
	month = {5/18/2024},
	pages = {1--56},
	title = {Large Language Models and the Argument from the Poverty of the Stimulus},
	year = {2024}}

@inproceedings{KalliniPapadimitriouFutrellMahowaldPotts:2024,
	 author    = {Julie Kallini and Isabel Papadimitriou and Richard Futrell and Kyle Mahowald and Christopher Potts},
  title     = {Mission: Impossible Language Models},
  booktitle = {Proceedings of the 62nd Annual Meeting of the Association for Computational Linguistics (Volume 1: Long Papers)},
  pages     = {14691--14714},
  year      = {2024},
  month     = aug,
  address   = {Bangkok, Thailand},
  publisher = {Association for Computational Linguistics},
  doi       = {10.18653/v1/2024.acl-long.787},
  url       = {https://aclanthology.org/2024.acl-long.787/}
}

@article{MahowaldIvanovaBlankKanwisherTenenbaumFedorenko:2024,
	author = {Kyle Mahowald and Anna A. Ivanova and Idan A. Blank and Nancy Kanwisher and Joshua B. Tenenbaum and Evelina Fedorenko},
	journal = {Trends in Cognitive Sciences},
	number = {6},
	pages = {517-540},
	title = {Dissociating language and thought in large language models},
	volume = {28},
	year = {2024}}

@article{Yngve:1960,
	author = {Yngve, Victor H.},
	journal = {Proceedings of the American Philosophical Society},
	number = {5},
	pages = {444--466},
	title = {A Model and an Hypothesis for Language Structure},
	volume = {104},
	year = {1960}}

@inproceedings{GulordavaBojanowskiGraveLinzenBaroni:2018,
	author = {Gulordava, Kristina and Bojanowski, Piotr and Grave, Edouard and Linzen, Tal and Baroni, Marco},
	booktitle = {Proceedings of NAACL 2018},
	pages = {1195--1205},
	title = {Colorless green recurrent networks dream hierarchically},
	year = {2018}}

@article{LinzenBaroni:2020,
	author = {Linzen, Tal and Baroni, Marco},
	journal = {Annual Reviews of Linguistics},
	title = {Syntactic Structure from Deep Learning},
	year = {2020}}

@incollection{Baroni:2022,
  author       = {Baroni, Marco},
  title        = {On the Proper Role of Linguistically-Oriented Deep Net Analysis in Linguistic Theorizing},
  booktitle    = {Algebraic Structures in Natural Language},
  editor       = {Shalom Lappin and J{\"u}rgen Bernardy},
  publisher    = {CRC Press / Taylor \& Francis},
  year         = {2022},
  pages        = {1--16},
  note         = {Preprint available as arXiv:2106.08694},
  doi          = {10.1201/9781003205388-1},
  url          = {https://arxiv.org/abs/2106.08694}
}

@inproceedings{Abdou:2022,
  author    = {Mostafa Abdou and Vinit Ravishankar and Artur Kulmizev and Anders S{\o}gaard},
  title     = {Word order does matter and shuffled language models know it},
  booktitle = {Proceedings of the 60th Annual Meeting of the Association for Computational Linguistics (Volume 1: Long Papers)},
  year      = {2022},
  pages     = {6907--6919},
  address   = {Dublin, Ireland},
  publisher = {Association for Computational Linguistics},
  doi       = {10.18653/v1/2022.acl-long.476},
  url       = {https://aclanthology.org/2022.acl-long.476},
}

@inproceedings{FongBerwick:2008,
  author    = {Sandiway Fong and Robert C. Berwick},
  title     = {Treebank Parsing and Knowledge of Language: A Cognitive Perspective},
  booktitle = {Proceedings of the 30th Annual Conference of the Cognitive Science Society},
  year      = {2008},
  pages     = {539},
  address   = {Washington, DC},
  publisher = {Cognitive Science Society},
  note      = {Paper No. 539},
}

@article{FutrellHahn:2025,
  title={Linguistic structure from a bottleneck on sequential information processing},
  author={Futrell, Richard and Hahn, Michael},
  journal={Nature Human Behaviour},
  year={2025},
  month={nov},
  day={24},
  publisher={Nature Publishing Group},
  doi={10.1038/s41562-025-02336-w},
  url={https://www.nature.com/articles/s41562-025-02336-w}
}

@article{Clark2023,
  author    = {Thomas Hikaru Clark and Clara Meister and Tiago Pimentel and Michael Hahn and Ryan Cotterell and Richard Futrell and Roger Levy},
  title     = {A Cross-Linguistic Pressure for Uniform Information Density in Word Order},
  journal   = {Transactions of the Association for Computational Linguistics},
  volume    = {11},
  pages     = {1048--1065},
  year      = {2023},
  publisher = {MIT Press},
  doi       = {10.1162/tacl_a_00589},
  url       = {https://aclanthology.org/2023.tacl-1.59/}
}

@incollection{Fong:2013,
  author    = {Sandiway Fong and Igor Malioutov and Beracah Yankama and Robert C. Berwick},
  title     = {Treebank Parsing and Knowledge of Language},
  booktitle = {Cognitive Aspects of Computational Language Acquisition},
  editor    = {Aline Villavicencio and Thierry Poibeau and Anna Korhonen and Afra Alishahi},
  series    = {Theory and Applications of Natural Language Processing},
  pages     = {133--172},
  publisher = {Springer},
  year      = {2013},
  doi       = {10.1007/978-3-642-35227-0_6},
}

@inproceedings{Vaswani:2017,
  author    = {Ashish Vaswani and Noam Shazeer and Niki Parmar and Jakob Uszkoreit and Llion Jones and Aidan N. Gomez and {\L}ukasz Kaiser and Illia Polosukhin},
  title     = {Attention Is All You Need},
  booktitle = {Advances in Neural Information Processing Systems (NeurIPS 2017)},
  year      = {2017},
  volume    = {30},
  pages     = {5998--6008},
  publisher = {Curran Associates, Inc.},
  url       = {https://papers.nips.cc/paper/7181-attention-is-all-you-need},
}

@article{Elman:1990,
  author    = {Elman, Jeffrey L.},
  title     = {Finding Structure in Time},
  journal   = {Cognitive Science},
  year      = {1990},
  volume    = {14},
  number    = {2},
  pages     = {179--211},
  doi       = {10.1207/s15516709cog1402_1},
}

@article{Musso:2003,
  author    = {Mariacristina Musso and Andrea Moro and Volkmar Glauche and Michel Rijntjes and Juergen Reichenbach and Christian B{\"u}chel and Cornelius Weiller},
  title     = {Broca's area and the language instinct},
  journal   = {Nature Neuroscience},
  year      = {2003},
  volume    = {6},
  number    = {7},
  pages     = {774--781},
  doi       = {10.1038/nn1077},
}

@article{Devlin:2018,
  author       = {Jacob Devlin and
                  Ming{-}Wei Chang and
                  Kenton Lee and
                  Kristina Toutanova},
  title        = {{BERT:} Pre-training of Deep Bidirectional Transformers for Language
                  Understanding},
  journal      = {CoRR},
  volume       = {abs/1810.04805},
  year         = {2018},
  eprinttype    = {arXiv},
  eprint       = {1810.04805},
  bibsource    = {dblp computer science bibliography, https://dblp.org}
}

@unpublished{ChemlaNefdt:2024,
	author = {Chemla, Emmanuel and Nefdt, Ryan},
	month = {July},
	note = {Ms.},
	title = {No Such Thing as a General Learner: Language models and their dual optimization},
	year = {2024}}

@article{FoxKatzir:2024,
	author = {Fox, Danny and Katzir, Roni},
	journal = {Theoretical Linguistics},
	number = {1--2},
	pages = {71--76},
	title = {Large Language Models and theoretical linguistics},
	volume = {50},
	year = {2024}}

@book{Chomsky:2000,
  author       = {Chomsky, Noam},
  title        = {New Horizons in the Study of Language and Mind},
  publisher    = {Cambridge University Press},
  year         = {2000},
  address      = {Cambridge; New York},
  isbn         = {0-521-65147-6},
  isbn13       = {978-0-521-65147-9},
  pages        = {xvii, 230},
  note         = {Hardback edition, with foreword by Neil Smith}
}

@article{HauserChomskyFitch:2002,
  author       = {Hauser, Marc D. and Chomsky, Noam and Fitch, W. Tecumseh},
  title        = {The Faculty of Language: What Is It, Who Has It, and How Did It Evolve?},
  journal      = {Science},
  year         = {2002},
  volume       = {298},
  number       = {5598},
  pages        = {1569--1579},
  doi          = {10.1126/science.298.5598.1569},
}

@misc{Xu:2025,
      title={Can Language Models Learn Typologically Implausible Languages?}, 
      author={Tianyang Xu and Tatsuki Kuribayashi and Yohei Oseki and Ryan Cotterell and Alex Warstadt},
      year={2025},
      eprint={2502.12317},
      archivePrefix={arXiv},
      primaryClass={cs.CL},
      url={https://arxiv.org/abs/2502.12317}, 
}

@article{Culbertson:2015,
  author    = {Jennifer Culbertson and Elissa L. Newport},
  title     = {Harmonic biases in child learners: In support of language universals},
  journal   = {Cognition},
  volume    = {139},
  pages     = {71--82},
  year      = {2015},
  doi       = {10.1016/j.cognition.2015.02.007},
}

@incollection{Collins:2016,
  author    = {Jeremy Collins},
  title     = {Some language universals are historical accidents},
  booktitle = {Explanation in Typology: Diachronic Sources, Functional Motivations and the Nature of the Evidence},
  editor    = {Karsten Schmidtke-Bode and Natalia Levshina and Susanne Maria Michaelis and Ilja A. Seržant},
  pages     = {51--65},
  publisher = {Language Science Press},
  year      = {2016},
  url       = {https://langsci-press.org/catalog/view/220/1524/1448-1},
}

@article{FutrellHahn:2022,
  author       = {Richard Futrell and Michael Hahn},
  title        = {Information Theory as a Bridge Between Language Function and Language Form},
  journal      = {Frontiers in Communication},
  volume       = {7},
  number       = {657725},
  year         = {2022},
  doi          = {10.3389/fcomm.2022.657725},
  url          = {https://www.frontiersin.org/articles/10.3389/fcomm.2022.657725/full}
}

@book{Chomsky1981,
  author    = {Noam Chomsky},
  title     = {Lectures on Government and Binding},
  year      = {1981},
  publisher = {Foris Publications},
  address   = {Dordrecht, Netherlands},
}

@inproceedings{Sinha:2021,
    title = "Masked Language Modeling and the Distributional Hypothesis: Order Word Matters Pre-training for Little",
    author = "Sinha, Koustuv  and
      Jia, Robin  and
      Hupkes, Dieuwke  and
      Pineau, Joelle  and
      Williams, Adina  and
      Kiela, Douwe",
    editor = "Moens, Marie-Francine  and
      Huang, Xuanjing  and
      Specia, Lucia  and
      Yih, Scott Wen-tau",
    booktitle = "Proceedings of the 2021 Conference on Empirical Methods in Natural Language Processing",
    month = nov,
    year = "2021",
    address = "Online and Punta Cana, Dominican Republic",
    publisher = "Association for Computational Linguistics",
    url = "https://aclanthology.org/2021.emnlp-main.230/",
    doi = "10.18653/v1/2021.emnlp-main.230",
    pages = "2888--2913",
}

@article{Futrell:2025,
  author       = {Futrell, Richard and Mahowald, Kyle},
  title        = {How Linguistics Learned to Stop Worrying and Love the Language Models},
  journal      = {Behavioral and Brain Sciences},
  year         = {2025},
  note         = {Accepted manuscript},
  doi          = {10.1017/S0140525X2510112X},
  url          = {https://arxiv.org/abs/2501.17047}
}

@inproceedings{ArnettBergen:2025,
  author    = {Catherine Arnett and Benjamin K. Bergen},
  title     = {Why do language models perform worse for morphologically complex languages?},
  booktitle = {Proceedings of the 31st International Conference on Computational Linguistics (COLING 2025)},
  pages     = {6607--6622},
  year      = {2025},
  publisher = {Association for Computational Linguistics},
  address   = {Mexico City, Mexico}
}

@unpublished{Milway:2025,
  author       = {Daniel Milway},
  title        = {On modern language models, impossible languages and anti-science},
  note         = {LingBuzz preprint. \url{https://ling.auf.net/lingbuzz/008314}},
  year         = {2025}
}

@article{Leivada:2025,
  author    = {Evelina Leivada and Raquel Montero and Paolo Morosi and Natalia Moskvina and Tamara Serrano and Marcel Aguilar and Fritz Guenther},
  title     = {Large Language Model probabilities cannot distinguish between possible and impossible language},
  journal   = {arXiv preprint},
  volume    = {arXiv:2509.15114},
  year      = {2025},
  url       = {https://arxiv.org/abs/2509.15114},
  note      = {Submitted on 18 Sep 2025}
}

@book{Moro:2016,
  author    = {Andrea Moro},
  title     = {Impossible Languages},
  publisher = {MIT Press},
  year      = {2016},
  address   = {Cambridge, MA},
  isbn      = {9780262034890},
}

@article{Baker:2012,
author = {Baker, Mark},
year = {2012},
month = {09},
pages = {},
title = {Formal Generative Typology},
isbn = {019954400X},
doi = {10.1093/oxfordhb/9780199544004.013.0012}
}

@inproceedings{yang2025anything,
    title = "Anything Goes? A Crosslinguistic Study of (Im)possible Language Learning in {LM}s",
    author = "Yang, Xiulin  and
      Aoyama, Tatsuya  and
      Yao, Yuekun  and
      Wilcox, Ethan",
    editor = "Che, Wanxiang  and
      Nabende, Joyce  and
      Shutova, Ekaterina  and
      Pilehvar, Mohammad Taher",
    booktitle = "Proceedings of the 63rd Annual Meeting of the Association for Computational Linguistics (Volume 1: Long Papers)",
    month = jul,
    year = "2025",
    address = "Vienna, Austria",
    publisher = "Association for Computational Linguistics",
    url = "https://aclanthology.org/2025.acl-long.1264/",
    doi = "10.18653/v1/2025.acl-long.1264",
    pages = "26058--26077",
    ISBN = "979-8-89176-251-0"
}

@Inbook{Schmid:1999,
author="Schmid, H.",
editor="Armstrong, Susan
and Church, Kenneth
and Isabelle, Pierre
and Manzi, Sandra
and Tzoukermann, Evelyne
and Yarowsky, David",
title="Improvements in Part-of-Speech Tagging with an Application to German",
bookTitle="Natural Language Processing Using Very Large Corpora",
year="1999",
publisher="Springer Netherlands",
address="Dordrecht",
pages="13--25",
isbn="978-94-017-2390-9",
doi="10.1007/978-94-017-2390-9_2",
url="https://doi.org/10.1007/978-94-017-2390-9_2"
}

@inproceedings{shmidman:2025,
    title = "Restoring Missing Spaces in Scraped {H}ebrew Social Media",
    author = "Shmidman, Avi  and
      Shmidman, Shaltiel",
    editor = "Bak, JinYeong  and
      Goot, Rob van der  and
      Jang, Hyeju  and
      Buaphet, Weerayut  and
      Ramponi, Alan  and
      Xu, Wei  and
      Ritter, Alan",
    booktitle = "Proceedings of the Tenth Workshop on Noisy and User-generated Text",
    month = may,
    year = "2025",
    address = "Albuquerque, New Mexico, USA",
    publisher = "Association for Computational Linguistics",
    url = "https://aclanthology.org/2025.wnut-1.3/",
    pages = "16--25",
    ISBN = "979-8-89176-232-9",
}

\section*{Appendix}
\appendix

\section{License Compliance}

In this study, we utilize datasets from the \hyperlink{https://github.com/facebookresearch/colorlessgreenRNNs}{Colorless Green RNNs} project, which are licensed under the Creative Commons Attribution–NonCommercial (CC BY-NC) license. Additionally, we adopt aspects of the methodology described in Kallini et al. (2024), whose \hyperlink{https://github.com/jkallini/mission-impossible-language-models}{code} is available under the MIT License. Both the datasets and the methodological framework are duly cited in the main text.

\section{Models used for POS tagging}

\begin{table}[h]
    \centering
    \begin{tabular}{c|c}
        \toprule
         Language & Model \\
         \midrule
         Danish & spaCy da\_core\_news\_md \\
         English & spaCy en\_core\_web\_sm \\
         Finnish & spaCy fi\_core\_news\_md \\
         French & 
         spaCy fr\_core\_news\_lg \\
         German & 
         spaCy de\_core\_news\_sm \\
         Greek & spaCy el\_core\_news\_md \\
         Hebrew & DictaBERT \citep{shmidman:2025} \\
         Italian & spaCy it\_core\_news\_sm \\ 
         Russian & spaCy ru\_core\_news\_sm \\
         
    \end{tabular}
    \caption{Models used for POS tagging in each experiment.}
    \label{tab:spaCy}
\end{table}

\end{document}